\documentclass[conference]{IEEEtran}
\IEEEoverridecommandlockouts
\usepackage{cite}
\usepackage{amsmath,amssymb,amsfonts}
\usepackage{algorithm,algorithmic}
\usepackage{graphicx}
\usepackage{textcomp}
\usepackage{siunitx}
\usepackage{tikz}
\usepackage{lipsum}
\usepackage[draft]{hyperref}

\usepackage{amsmath}
\usepackage{amsfonts}
\usepackage{amssymb}
\usepackage{mathtools}

\usepackage{algorithm}

\def\BibTeX{{\rm B\kern-.05em{\sc i\kern-.025em b}\kern-.08em
T\kern-.1667em\lower.7ex\hbox{E}\kern-.125emX}}

\newcommand\copyrighttext{%
  \footnotesize \textcopyright 2018 IEEE. Personal use of this material is permitted. Permission from IEEE must be obtained for all other uses, in any current or future media, including reprinting/republishing this material for advertising or promotional purposes, creating new collective works, for resale or redistribution to servers or lists, or reuse of any copyrighted component of this work in other works.
  DOI: \href{}{}}
\newcommand\copyrightnotice{%
\begin{tikzpicture}[remember picture,overlay]
\node[anchor=south,yshift=10pt] at (current page.south) {\fbox{\parbox{\dimexpr\textwidth-\fboxsep-\fboxrule\relax}{\copyrighttext}}};
\end{tikzpicture}%
}

\begin{document}




\title{Using a Game Engine to Simulate Critical Incidents and Data Collection by Autonomous Drones\\
\thanks{This research has received funding from the European Union's Horizon 2020 Programme under Grant Agreement No. 700264.}
}
\author{\IEEEauthorblockN{David L. Smyth}
\IEEEauthorblockA{\textit{School of Computer Science} \\
\textit{National University of Ireland Galway}\\
Galway, Ireland \\
d.smyth10@nuigalway.ie}
\and
\IEEEauthorblockN{Frank G. Glavin}
\IEEEauthorblockA{\textit{School of Computer Science} \\
\textit{National University of Ireland Galway}\\
Galway, Ireland \\
frank.glavin@nuigalway.ie}
\and
\IEEEauthorblockN{Michael G. Madden}
\IEEEauthorblockA{\textit{School of Computer Science} \\
%
\textit{National University of Ireland Galway}\\
Galway, Ireland \\
michael.madden@nuigalway.ie}
}
\maketitle
\copyrightnotice
\begin{abstract}
Using a game engine, we have developed a virtual environment which models important aspects of critical incident scenarios. We focused on modelling phenomena relating to the identification and gathering of key forensic evidence, in order to develop and test a system which can handle chemical, biological, radiological/nuclear or explosive (CBRNe) events autonomously. This  allows us to build and validate AI-based technologies, which can be trained and tested in our custom virtual environment before being deployed in real-world scenarios. We have used our virtual scenario to rapidly prototype a system which can use simulated Remote Aerial Vehicles (RAVs) to gather images from the environment for the purpose of mapping. Our environment provides us with an effective medium through which we can develop and test various AI methodologies for critical incident scene assessment, in a safe and controlled manner.
%
\end{abstract}

\section{Motivation and Context}
Accidents and attacks that involve chemical, biological, radiological/nuclear or explosive (CBRNe) substances are rare, but can be of high consequence due to the need for forensic investigators and other personnel to face unquantified threats. These events are exceedingly difficult to prepare for because they are so rare and diverse that limited data from real events is available. Physical training exercises are not frequently held due to cost, which also limits the amount of simulated data that is available. A consequence of this is that any systems that are developed to provide support can be difficult to evaluate and validate. To address this, our goal is to design and develop high-fidelity simulated environments, which allow us to maintain the core aspects of CBRNe incidents without any risk of exposure to the dangerous elements of such scenes in the real world. 

\subsection{Related Research}
Virtual environments designed using games engines have previously been used to gather data to train models for a range of applications \cite{1608.02192}\cite{uav_benchmark_simulator} and software packages have been written that can generate photo realistic images from games engines\cite{1609.01326}. There have also been attempts made to develop models of physical systems that simulate potentially dangerous environments \cite{4625089}. To the best of our knowledge, there are no prior applications using games engines to model critical incidents for the purpose of developing analytical tools in a virtual setting that will then transfer to real-world deployment.

\subsection{Contributions}
Our primary contribution is the creation of a simulated critical incident scenario using Unreal Engine 4, UE4 (See Section II). We developed this to be easy to modify, extensible, and compatible with real-world robotic and sensor systems. We have  fully integrated the \emph{AirSim} \cite{airsim2017fsr} plugin into the environment to provide Remote Aerial Vehicles (RAVs) and Remote Ground Vehicles (RGVs) with access to modified APIs which can be used to carry out tasks, using GPS coordinates as opposed to the internal UE4 North, East, Down (NED) coordinate system (Section III). In addition, we have developed a baseline system for autonomously gathering images in a specified region in the environment using multiple RAVs (Section IV). The virtual environment is freely available on GitHub for use by the research community (See: \href{https://github.com/ROCSAFE/CBRNeVirtualEnvMultiRobot/releases}{ROCSAFE CBRNe environment}\footnote{\href {https://github.com/ROCSAFE/CBRNeVirtualEnvMultiRobot/releases}{https://github.com/ROCSAFE/CBRNeVirtualEnvMultiRobot/releases}}).

\section{Virtual Environment Design}
Within the context of a project called ROCSAFE (Remotely Operated CBRNe Scene Assessment \& Forensic Examination)\footnote{\href {http://www.rocsafe.eu}{http://www.rocsafe.eu}}, several operational scenarios have been identified to represent distinct classes of CBRNe threats. We are developing a virtual representation of these scenarios to validate various AI technologies. Our initial environment consists of a rural scene with a train transporting radioactive material that has been de-railed. 
\subsection{Unreal Engine 4}
We selected UE4 to develop the environment for reasons which include: UE4 can produce near-photo-realistic images; there are several plugins for UE4 that provide virtual RAVs/RGVs; and UE4's blueprint visual scripting system (described in the next section) can allow for rapid prototyping. 

We identified the assets that would be necessary to develop the environment to our specification. These assets were acquired from websites such as 
\href{http://www.cgtrader.com}{CGTrader}\footnote{\href {http://www.cgtrader.com}{https://www.cgtrader.com}}
and 
\href{https://3dwarehouse.sketchup.com/?hl=en}{Google 3D warehouse}\footnote{\href {https://3dwarehouse.sketchup.com/?hl=en}{https://www.3dwarehouse.sketchup.com}}. The assets were imported into \emph{Autodesk 3DS} in order to ensure that textures were of sufficient quality to facilitate the planned image processing on collected images. They were then 
imported into UE4 as static meshes and placed into the scene. This process was designed to be scalable and static meshes can be replaced by simply importing a new mesh which will retain the position of the original in the environment. 

\subsection{Simulating Radiation using Blueprints}
UE4 supports a visual scripting system known as \textit{blueprints}, which is a game-play scripting system based on the concept of using a node-based interface to create game-play elements from within the UE4 Editor. A blueprint was created to simulate gamma radiation in the environment. First, we developed a model of the level of ionizing radiation observed $d$ meters from a source, making the assumptions that the radiation is emitted from a point and is transmitted equally in all directions. We also assumed that the propagating medium is lossless, implying that the level of ionizing radiation is conserved. Letting $d$ = distance from point source of radiation, the strength of the ionizing radiation in micro Sieverts per second (\si{\micro\sievert\per\second}) is determined by the equation  $strength(d) \propto \frac{1}{d^2}$. Letting the strength at a distance of 1m from the source equal to
\sisetup{quotient-mode=fraction}
$\sigma$\si{\micro\sievert\per\second}
, the ionizing radiation in \si{\micro\sievert\per\second} is given by:
\[strength(d) = \frac{\sigma \times 4 \times \pi}{d^2 \times 4 \times \pi} = \frac{\sigma}{d^2} \si{\micro\sievert\per\second}\]

This was modelled in UE4 by using a built-in blueprint node, 
\href{https://api.unrealengine.com/INT/API/Runtime/Engine/Kismet/UGameplayStatics/ApplyRadialDamageWithFalloff/index.html}{\emph{Apply Radial Damage with Falloff}}\footnote{\href {https://api.unrealengine.com/INT/API/Runtime/Engine/Kismet/UGameplayStatics/ApplyRadialDamageWithFalloff/index.html}{https://api.unrealengine.com}}.
The detection of ionizing radiation strength can then be simulated using the
\href{https://docs.unrealengine.com/en-us/Engine/Blueprints/UserGuide/Events/\#eventradialdamage}{\emph{Event Radial Damage}}
\footnote{\href {https://docs.unrealengine.com/en-us/Engine/Blueprints/UserGuide/Events/\#eventradialdamage}{https://docs.unrealengine.com}} node. This blueprint can be integrated with external plugins that facilitate the control of vehicles in the environment and can be exposed via an API relating to the vehicle. Similarly, it is possible to develop models of chemical and biological dispersion with a corresponding simulated detection. This allows us to decouple the development of the decision support analysis upon detection of hazardous substances from the actual details of the sensors developed in the project, streamlining the development of AI systems.

\section{Integration and Extension of AirSim}\label{sec:AirSim}
In order to develop autonomous routing algorithms for RAVs and RGVs, we used the AirSim \cite{airsim2017fsr} plugin for UE4. We chose to use AirSim over other plugins for the following reasons: 
\begin{itemize}
\item AirSim has APIs for C++, Python and C\# which allows flexibility in the language used to control vehicles.
\item AirSim provides the ability to run multiple RAVs in the same environment.
\item AirSim has been open-sourced with an MIT licence and was designed to be easily extended.
\item The AirSim API provides convenient access to unmodified images, depth maps and segmented images.
\end{itemize}
AirSim internally models RAVs using different low-level modules outlined in \cite{airsim2017fsr} and exposes the RAV as a blueprint that we extended to sense radiation placed in the environment and to provide debugging lines for the drone flight path, as shown in Figure \ref{fig:agentRoutes}. 

\subsection{Adding GPS Support}
The AirSim API provides support to query the RAV GPS position using the WGS 84 coordinate system, where the home GPS location of the RAV is set arbitrarily.
The API only provides the ability to send the RAV to an Unreal North, East, Down (NED) position relative to the RAV home position, set to (0,0,0) in the UE4 NED coordinate system. We added the capability to set the RAV home GPS position and send the RAV to GPS coordinates rather than NED coordinates. We developed this by creating a GPS coordinate class in the Python API, with methods to calculate the difference between one GPS point and another in terms of degrees, which could in turn be converted to the number of metres latitude/longitude needed to be travelled to reach the second GPS point. Finally, using the UE4 conversion rate of 1 Unreal Unit to 1 cm, we calculated the NED transformation needed to send the RAV to the second GPS point, which allowed us to use functionality already provided by the API.
\\

\begin{figure}[h]
\centering
\includegraphics[width=3.2in]{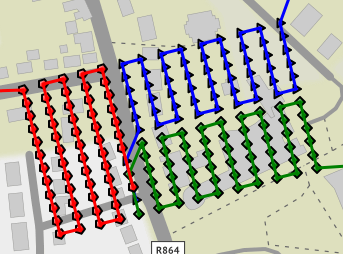}
\caption{Planned routes for RAVs plotted on a map for validation purposes.}
\label{fig:agentRoutesLeaflet}
\end{figure}

\begin{figure}[h]
\centering
\includegraphics[width=3.2in]{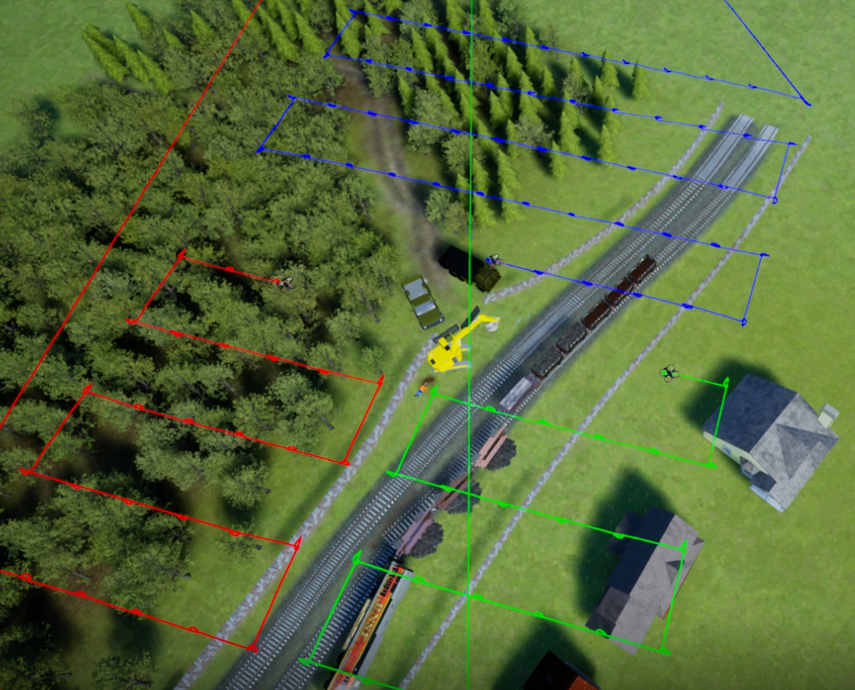}
\caption{The virtual environment with RAVs visiting calculated waypoints}
\label{fig:agentRoutes}
\end{figure}

Figure \ref{fig:agentRoutesLeaflet} shows the planned routes of three RAVs overlaid on a known location (The National University of Ireland, Galway) which was used to validate the route planning algorithms.
Figure \ref{fig:agentRoutes} shows three RAVs in our developed Unreal Engine simulation environment following the planned routes to gather simulated data from the environment.

\subsection{Collecting Images in the Virtual Environment}\label{sec:algodesign}
The first problem we tackled using our virtual environment was collecting images to provide an overview of the scene. This task naturally breaks down into the following sub-tasks: 
\begin{itemize}
\item Identify the region that needs to be mapped.

\item Divide the region into discrete tiles, the centre of which will define the point where an image will need to be taken to cover that tile. The spacing between tiles should provide approximately a 20\% overlap between the images recorded for each tile to provide optimal image stitching results using algorithms such as ORB \cite{6126544}.
\item Design an algorithm that will assign a set of points to each RAV. These sets should form a partition of the full set of points. The objective should be to minimize the time taken for the longest flight path.

\item For each waypoint of the calculated RAV routes, collect an image with GPS position and camera metadata and send each collected image to a repository for analysis.
\end{itemize}
These tasks are subject to the assumptions that the RAVs operate at a fixed height which was chosen to avoid possible obstacles (32m), that the desired region to map is a regular polygonal region specified using the WGS 84 coordinate system and that the RAVs operate at a known average velocity for the duration of their flight. It is also assumed that the region in which the RAVs will operate in is small enough to be considered a flat plane 32m above the earth's surface.

\section{Generation of Waypoints and Routes for RAVs}
In this context, waypoints are GPS coordinates that RAVs travel to in order to collect data, and a sequence of waypoints defines a route.
From our work in extending the API of the AirSim plugin, we were able to generate waypoints for RAVs using a GPS coordinate class that could be mapped to UE4 coordinates. This means that we will be able to extend the work outlined in this paper to RAVs that do not operate in a virtual environment. The method for generating the waypoints is as follows, in the context of gathering data related to the simulated CBRNe scene spanning a polygonal region: 
\begin{enumerate}
\item Among the bounding coordinates which define the polygon, identify the highest latitude, highest longitude, lowest latitude, lowest longitude. 
Generate a circum-rectangle from these coordinates which contains the polygon. The corner values defining the circum-rectangle are treated as points on the $x,y$ plane, forming their own finite rectangular plane as per the assumption outlined in \ref{sec:algodesign}. 

\item \label{point2} For the task of mapping, the desired width, $w$ between the points at which images are recorded for an overlap of $y \times 100\%$ between images, using a camera with a field of view (FOV) of $2 \times \theta$ degrees, at a height of $h$ meters, is given by the equation $w=2 \times h \times \frac{1-y}{y+1} \times \tan(\theta)$.
We allow grid tiles to lie partially outside the grid to facilitate the full mapping of the grid. 

\item Calculate the size and number of increments (using \ref{point2}).) necessary to generate waypoints in both directions specified by the axes of the circum-rectangle. The (i, j) waypoint is calculated by taking the sum of the coordinate specified by the lowest latitude, lowest longitude and the increment along the first axis of the rectangle multiplied $i$ times plus the increment along the second axis of the rectangle multiplied $j$ times to give the next waypoint position.
\item In order to remove the waypoints that lie inside the circum-rectangle but not the polygon, perform the following test to determine whether or not to delete the point: 
\\
For the given point to be tested, extend a ray from the point horizontally and record the number of times that it crosses the boundary of the polygon. If the line extended from the point being tested enters the polygon, it must exit again (since it is of infinite length and the region is finite), which means that an even number of boundary crossings implies that the point lies outside the polygon and an odd number implies that the point lies inside the polygon. This is a well known-algorithm, with documented implementations \cite{Shimrat:1962:APP:368637.368653}.
\end{enumerate}

Each RAV in the environment is treated as an agent, together forming a multi-agent system. These agents are treated as heterogeneous, as some RAVs may have sensors and other equipment attached to perform a specific task. Each agent is assumed to share the same internal representation of the environment, namely the connected graph defined by the generated waypoints, where each point is initially unvisited. In order to gather data at each of the waypoints, the route generating algorithm needs to satisfy the following: given a set $P$ of points to visit and $n$ RAV agents, find a subtour for each agent such that each point in $P$ is visited exactly once by any agent in the system, with the objective of minimizing the longest time taken for any individual agent subtour, in order to minimize the time taken to carry out the mapping task. This follows from the fact that the time taken to complete the overall mapping task equates to the longest time taken for any agent to complete their subtour. This is the Multiple Traveling Salesperson problem \cite{multipleTSP}. 

The approach taken was a centralized one, with the intention that we extend the work completed to a distributed form of the problem, where agents are not necessarily informed of a plan from a centralized source. A nearset neighbor algorithm was implemented as follows:







\begin{algorithm}
\caption{Generating the RAV Agent Routes}
\label{alg:agentRoutesEdited}
\begin{algorithmic}[1]
\renewcommand{\algorithmicrequire}{\textbf{Input:}}
\renewcommand{\algorithmicensure}{\textbf{Output:}}
\REQUIRE Array of agents, Set of coordinates, Cost function which takes two GPS coordinates and outputs a real number representing the cost of travelling from one to the other. 
\ENSURE  A key-value container of agents and their corresponding routes.\\
\hfill\pagebreak

\noindent\textbf{\textit{\noindent Initialization} :}\\
agentPaths$\leftarrow$empty key-value container\\
visitedPoints$\leftarrow$empty array
\\
For each agent in agents:\\
\quad Initialise path of agent as empty array in agentPaths
currentAgentIndex$\leftarrow$0\\
currentAgent$\leftarrow$agents.get(currentAgentIndex)\\
\hfill\pagebreak

\WHILE {pointsToVisit is not empty}

\STATE agentPosition $\leftarrow$ last value in agentPaths.get(agent)

\STATE P$\leftarrow$\(\displaystyle \min_{p \in pointsToVisit}\)cost(agentPosition, p)

\STATE Update currentAgent value in AgentPaths to include P
\STATE Add P to visitedPoints.
\STATE Remove P from pointsToVisit.
\STATE currentAgentIndex$\leftarrow$(currentAgentIndex+1) $\mathbf{mod}$$\vert$List of Agents$\vert$
\STATE currentAgent$\leftarrow$agents.get(currentAgentIndex)

\ENDWHILE
\RETURN agentRouteMap
\end{algorithmic} 
\end{algorithm}

\subsection{Algorithm Motivation}
In designing this algorithm, we wanted to ensure that each RAV has a similar amount of work to perform in order to balance the work to be done across the system. The  generated grid of points to visit is evenly spaced, which means that the cost in terms of distance of travelling to adjacent grid points is fixed along the two dimensions defined by the circum-rectangle of the polygonal region. This means our algorithm encourages RAVs to add the same (minimum) cost of travelling to an adjacent node to their cost each time they choose an additional node to visit, usually resulting in a balanced distribution of work across the system.

\subsection{Algorithm  Performance Evaluation}
The algorithms performance is measured by finding the time taken for all of the generated routes to be completed, assuming that they are all started at the same time and all agents run in parallel. This equates to finding the distance of the longest generated path, assuming that all agents move at the same constant velocity.
We calculated a lower bound for the cost of running this algorithm for $n$ agents, found to be $\frac{1}{n}\times$ cost of optimal path for a single agent.
\\
\\Proof:\\
Let T be a tour that is an optimal solution to the Travelling Salesperson Problem (TSP) for  a graph G(V, E), where E $\subseteq$ V $\times$ V, with a cost function $c(p_i, p_j)$ defined for all $(p_i, p_j)$ $\in$ E and an induced cost function 
C(T) = $\sum\limits_{(p_i, p_j)\in S_k}C(p_i, p_j)$ defined for any tour of G.
The tour T is an ordered tuple (($p_i, p_j$), ($p_j, p_k$),..., ($p_m, p_n$)) $\subseteq$ E which satisfies Cost(T) = \(\displaystyle \min_{e \subseteq E}\) $\sum_{(p_i, p_j) \in e}$ c($p_i, p_j$) with the constraint that each $p_i \in$ V must be visited exactly once. This means T is a hamiltonian tour of G of minimal cost. Any partition ($T_1, T_2, ..., T_m$) of a solution T' can be written as C(T')=
\\

$\sum\limits_{k=1}^{m}\sum\limits_{(p_i, p_j)\in T_k} c(p_i, p_j) \leq m\times$
\(\displaystyle\max_{T_k \in T}\)
$\sum\limits_{(p_i, p_j)\in T_k}c(p_i, p_j)$
\\

\noindent and since Cost(T) $\leq$ Cost(T'), for the partition determined by Algorithm \ref{alg:agentRoutesEdited}, we find that for any solution S to the Multiple Traveling Salesperson (MTS) problem consisting of the partition ($S_1, S_2, ...,S_m$) for each of the m agents:\\
Cost(T)
$\leq$ Cost(S) $\leq$  m$\times$
(\(\displaystyle\max_{S_k \in S}\)
$\sum\limits_{(p_i, p_j)\in S_k}C(p_i, p_j)$) = Cost of MTS solution, S.
This means we can use a solution of the standard Traveling Salesperson problem as a lower bound for comparison with Algorithm \ref{alg:agentRoutesEdited}. For example, the Held-Karp algorithm gives a well-known lower bound when dealing with a metric space \cite{VALENZUELA1997157}.






\section{Conclusion}
In this paper, we have outlined the core steps taken to create a customized virtual environment using a games engine and the real-world applications that can benefit from its use. We use a modular approach in designing the virtual environment which provides extensibility and flexibility in our model of the world. This will be valuable once the world starts to become large in terms of the number of actors included. We outlined the process of development of the environment, which was carried out \emph{independently} of the plugins used in order to avoid dependencies that could prevent us from using different or updated plugins at a later stage.

We have also discussed our use of AirSim to provide us with a remote vehicle testbed to virtually test real-world scenarios in a safe and robust way. We extended the AirSim API to allow the querying of GPS position and the ability to send an RAV to a calculated GPS coordinate instead of an UE4 coordinate. A further enhancement was to include the use of debug lines to track the paths of each of the RAVs.

\subsection{Virtual Environments: Advantages vs Drawbacks}
As part of this research, we have considered the advantages  and drawbacks of using a virtual environment to test the scene assessment technologies which are being developed in parallel. Advantages of using a virtual environment include: 
\begin{itemize}
\item There is a low monetary cost associated with the development and many assets are freely available.
\item Virtual environments are highly configurable and updating realistic scene components is straightforward. This allows flexibility in the specification of the environment.
\item The virtual environment scales, whereas real-world events are limited by physical space. It is possible to add hundreds of actors to a scene without impacting engine performance significantly whereas in real life, producing a scene with hundreds of components is not feasible.
\item Events that are relatively slow to occur in the real world can be simulated in a games engine more rapidly by accelerating the clock. This allows for rapid gathering of data for analysis than may be possible in the real world.
\item The games engine provides a ground truth for everything included in the environment. This is valuable for image processing tasks such as automatically labelling data.
\end{itemize}
Drawbacks of using a virtual environment include:
\begin{itemize}
\item Simplified models of physical phenomena must be modelled in the virtual environment, which is time consuming and may not be fully accurate.
\item High-fidelity models, components, textures, and layouts are needed in order to achieve a virtual environment that can provide a solid basis for transfer of expected test results to the real world.
\item As noted in \cite{1609.01326}, training a computer vision model with images generated from a virtual environment may not consistently perform well on real-world images.

\end{itemize}

\subsection{Future Work}
We plan to design further virtual scenes using the same modular approach outlined in this paper in order to be able to test AI systems across a range of operational scenarios. This will give us the ability to validate these systems prior to field tests that are expensive to run. 

We also intend to build on our current work to develop more efficient routing algorithms and improved image stitching results. 
We plan to introduce inter-agent communication protocols to introduce redundancy into the system in case an agent goes offline. A decentralized approach will be taken for the routing algorithms which will be able to make real-time adjustments in case of the failure of one of the RAVs. We will also develop algorithms that will autonomously identify sources of hazardous material based on sensor readings. This will then
feed into a routing algorithm for ground vehicles to autonomously navigate to identified evidence collection points. Finally, we will refine the simulated environment by modifying textures and designing and locating more physically realistic assets to generate a large, annotated photo-realistic data set that can be used for computer vision applications such as pixel-level image annotation and object identification. 


\bibliographystyle{ieeetr} 
\bibliography{main} 

\end{document}